\documentclass[conference]{IEEEtran}
\IEEEoverridecommandlockouts
\usepackage{cite}
\usepackage{mathtools}
\usepackage{amsmath,amssymb,amsfonts}
\usepackage[noend]{algorithmic}

\usepackage{graphicx}
\usepackage{textcomp}
\usepackage{xcolor}

\usepackage{standalone}
\usepackage{algorithm}
\usepackage{todonotes}
\usepackage[normalem]{ulem}
\usepackage{enumitem}
\usepackage{url}
\usepackage{hyperref}

\def\BibTeX{{\rm B\kern-.05em{\sc i\kern-.025em b}\kern-.08em
    T\kern-.1667em\lower.7ex\hbox{E}\kern-.125emX}}
\begin{document}

\title{Federated Transfer Learning for Medical Image Classification under Mixed Skew\\
{\footnotesize Presented at 4th IEEE International Conference on Federated Learning Technologies and Applications (\href{https://flta-conference.org/flta-2026/}{FLTA 2026})}
\thanks{Région Normandie}
}

\author{
	\IEEEauthorblockN{Alexandre G. Leclercq\IEEEauthorrefmark{1}\IEEEauthorrefmark{2}\IEEEauthorrefmark{3}, Sébastien Bougleux\IEEEauthorrefmark{1}, Aurélien Corroyer-Dulmont\IEEEauthorrefmark{2}\IEEEauthorrefmark{3}, Romain Hérault\IEEEauthorrefmark{1}}
	
\IEEEauthorblockA{\IEEEauthorrefmark{1}\textit{Université Caen Normandie, ENSICAEN, CNRS, Normandie Univ, GREYC UMR 6072},
F-14000 Caen, France},
\IEEEauthorblockA{\IEEEauthorrefmark{2}\textit{Université de Caen Normandie, CNRS, Normandie Univ, ISTCT UMR6030, CYCERON}, 
	F-14000 Caen, France}
\IEEEauthorblockA{\IEEEauthorrefmark{3}\textit{Artificial Intelligence Department, Centre François Baclesse}, 
	 F-14000 Caen, France \\
	Email: \{alexandre.leclercq, sebastien.bougleux, romain.herault\}@unicaen.fr } a.corroyer-dulmont@baclesse.unicancer.fr
}

\maketitle

\begin{abstract}
Deep learning in medical imaging is constrained by data scarcity and by privacy
regulations that make centralizing patient data across institutions impractical.
Federated Learning (FL) addresses these constraints by keeping data local, but
its distributed nature exposes training to strong statistical heterogeneity.
Multicentric medical data is particularly affected, exhibiting a mixed skew that
combines feature, label, concept, and quantity shifts across centers.
Personalized FL mitigates heterogeneity by tailoring a model to each client, yet
most methods assume training from scratch, which is unrealistic given the high
dimensionality of medical features relative to the limited data available. We
propose the Personalized Aligner Federated Pretrained (PAFP) framework, a
personalized FL approach built on a pre-trained backbone. PAFP inserts
client-specific Feature Alignment modules before the fine-tuned blocks to realign
hidden feature distributions toward a server-aggregated target. On the FedISIC benchmark,
PAFP attains the highest unweighted balanced accuracy while reducing inter-client
performance variance roughly six-fold relative to the strongest baseline,
yielding accuracy and cross-client equity jointly. Our code is publicly available at \url{https://github.com/Alexandre-Leclercq/PAFP}.
\end{abstract}


\section{Introduction}
Machine learning and deep learning models are attracting increasing interest in the medical imaging field to address various tasks such as tumor segmentation, disease prediction, anomaly detection, and modality to modality translation \cite{LITJENS2017,zhou21,moor23}. 
However, one of the biggest problems that the community is facing when it comes to training a medical imaging deep learning model is the lack of data. This data scarcity is due to several factors \cite{zhou21,moor23}, such as the need for specialized expertise to annotate training data, the necessity of patient consent, and strict privacy regulations and legal barriers (GDPR in Europe, HIPAA in the USA, etc.) that protect patient data but make the process of sharing across institutions complicated, if not impossible.

Federated Learning (FL) is a machine learning paradigm that addresses these privacy regulations and legal barriers \cite{mcmahan17,kairouz21}. 
Instead of aggregating data from multiple clients (e.g., medical centers), as a traditional centralized learning approach would, FL keeps the data localized at every client and transmits the models to a server, whose goal is to gather and aggregate each client's model.
In particular, FL in medical settings is typically framed as a cross-silo scenario \cite{rieke20}. This implies a small number of clients, each holding a large dataset and sufficient computational power, and acting as a trustworthy party. This stands in contrast to cross-device FL, which assumes the opposite.
However, the distributed nature of FL introduces new challenges, particularly the issue of \textit{data heterogeneity} \cite{kairouz21,zhao2018,lu24}, where the data distribution is highly variable across clients (i.e., non-IID data). This problem creates diverging optimization directions during local training, which leads to a consensus issue commonly known as client drift. Medical imaging is particularly exposed to such phenomena, with feature skew, label skew, concept skew, and quantity skew naturally occurring across institutional datasets.
The objective of traditional FL is to learn a single global model that generalizes well across all client datasets. Such an objective can be very hard to achieve under severe heterogeneity. To overcome this limitation, Personalized Federated Learning (pFL) shifts the paradigm \cite{tan23}. In contrast to traditional FL, pFL does not attempt to learn a one-size-fits-all global model; instead, it aims to adapt and tailor models to the specific data distribution of each individual client. In this context, the global model essentially serves as a shared baseline to facilitate knowledge transfer toward the local clients.
However, most of the approaches proposed in the literature assume training from scratch. These methods are not realistically suitable for medical imaging applications, given that the feature space is highly dimensional while the amount of available training data is severely limited. Thus, it is necessary to leverage pre-trained models \cite{tajbakhsh16,raghu19,chen20,nguyen23} in this context.
In this work, we focus on the medical image classification task and validate our approach on the FedISIC2019 dataset from the FLamby benchmark \cite{NEURIPS2022_232eee8e}, a real-world, naturally partitioned federated skin-lesion classification problem that exhibits strong inter-client heterogeneity.
Our contributions are summarized as follows:
\begin{itemize}[leftmargin=*]
    \item We formalize the pFL problem in the context of medical imaging, a formulation whose scope extends beyond this domain
    \item We analyze the data shift across clients.
    \item We propose the Personalized Aligner Federated Pretrained (PAFP) framework, a pFL approach based on a pre-trained model, using injected alignment modules before the fine-tuned block, which is client specific.
\end{itemize}
    
\section{Related Works}

\textbf{Traditional FL} algorithms, such as FedAvg \cite{mcmahan17}, optimize a single global model under the implicit assumption of independent and identically distributed (IID) data across clients. Under realistic non-IID settings, statistical heterogeneity causes the global model to converge toward a sub-optimal compromise that may perform poorly on individual client distributions \cite{zhao2018}. To mitigate this client drift, methods like FedProx \cite{li20}, SCAFFOLD \cite{karimireddy20}, and MOON \cite{li21} introduce proximal terms, control variates, or contrastive losses, respectively. However, these optimization-based strategies still fundamentally aim to produce a single shared model, limiting their efficacy on highly skewed local data. 

\textbf{Personalized Federated Learning (pFL)} shifts the objective to a family of models tailored to local distributions. Local adaptation refines a global model either through simple fine-tuning or meta-learning initializations (e.g., Per-FedAvg \cite{fallah20}) for rapid convergence. Regularization-based methods like Ditto \cite{li21ditto} jointly optimize global and local models to balance consensus and personalization. Architecture-based methods introduce structural decoupling: FedBN \cite{li2021fedbn}, FedPer \cite{arivazhagan2019}, and FedRep \cite{collins21} retain specific local layers while sharing the feature extractor. Feature-decoupling methods (DBE \cite{zhang2023eliminating}, FDSE \cite{wang25dse}) extract client-invariant global features alongside localized representations. Finally, to specifically handle label skew, methods like FedLC \cite{zhang22} calibrate logits based on class probabilities to prevent overfitting to minority classes.

\textbf{Mixed Skew and Decoupled Personalization}. In practical medical scenarios, clients simultaneously face domain shift (sensor variability / feature skew) and class imbalance (pathology prevalence / label skew). This \textit{mixed skew} defeats traditional pFL methods designed for a single skew type, prompting the recent shift toward decoupling visual and statistical variability treatments. For instance, FED-ROD \cite{chen22} introduces a two-predictor framework on top of a shared, actively trained feature extractor. It optimizes a generic head using a class-balanced loss while learning a personalized add-on head via empirical risk minimization. Similarly, Fed-CO$_2$ \cite{cai2023fedco2} relies on mutual knowledge distillation between a global online model (using personalized BN layers) and a client-specific offline model. However, because these architectures require training the entire backbone from scratch or maintaining compute-intensive dual networks, they remain heavy and prone to overfitting or destabilization under severe data scarcity.

\textbf{Federated Transfer Learning (FTL) \& Parameter-Efficient Fine-Tuning (FedPEFT).} 
To address data scarcity in domains like medical imaging, FTL \cite{chen20} leverages pre-trained models to initialize the federated network.
However, naive fine-tuning across clients under severe non-IID conditions often causes catastrophic forgetting of the pre-trained prior or fails to properly personalize features \cite{nguyen23}. To achieve stable personalization without the overhead of full-network training, recent paradigms shift toward \textbf{FedPEFT}. 
Pioneered in NLP \cite{zhao23, zhang2023fedpetuning}, FedPEFT inserts lightweight adapters into a strictly frozen backbone. For example, AdaFL \cite{cai23} freezes over 99\% of parameters to align biased representations with negligible overhead.
This paradigm has transitioned to Computer Vision \cite{guo24, yang23pFedPG} and medical imaging (e.g., FedFMS \cite{liu24}, which federates the Segment Anything Model). Beyond simple insertion, recent works like PERADA \cite{xie24perada} regularize personalized adapters with a distilled global adapter, effectively balancing local personalization with global generalization capabilities. 
To address memory constraints on heterogeneous edge devices, frameworks like Fed-HeLLo \cite{zhang2025fedhello} propose Heterogeneous LoRA Allocation (HLA) to dynamically distribute trainable layers. Finally, for black-box foundation models, ZOOPFL \cite{lu24zoo} employs Zeroth-Order Optimization (ZOO) and input surgery to bypass internal backpropagation, while post-hoc methods like CCVR \cite{luo21nofear} debias the global classifier via virtual Gaussian representations to mitigate label skew.

\textbf{Robustness in Federated PEFT.} While standard FedPEFT methods focus on efficiency, they remain highly susceptible to noisy data and aggregation inconsistencies. Specifically, Low-Rank Adaptation (LoRA) components can inadvertently amplify bias in the presence of label noise \cite{yang23pFedPG}. To combat this, RFedLR \cite{fang2025rfedlr} proposes Sensitivity-aware Robust Tuning (SRT), which selectively updates noise-sensitive parameters based on gradient discrepancies between clean and noisy proxies. Furthermore, to address the sub-optimality of naive LoRA averaging, RFedLR introduces Adaptive Federated LoRA Aggregation (AFLA), which dynamically weights client updates based on parameter importance (measured via Fisher Information and noise sensitivity), rather than relying on simple parameter averaging.

\textbf{Positioning our approach.} While existing FedPEFT methods successfully reduce parameter overhead, they largely focus on generic personalization or global-local regularization (e.g., PERADA leverages knowledge distillation to reconcile global-local objectives). Frameworks such as RFedLR have highlighted the vulnerability of standard adapter-based methods to label noise.
However, these robust frameworks are primarily reactive, focusing on filtering or mitigating the impact of noisy inputs, rather than addressing the structural \textit{mixed skew} inherent in medical imaging. In contrast, PAFP adopts a proactive stance against distribution shift
by positioning internal alignment modules to standardize features \textit{before} the fine-tuned blocks. 

\begin{figure*}[!t]
    \centering
    \includegraphics[width=0.8\textwidth]{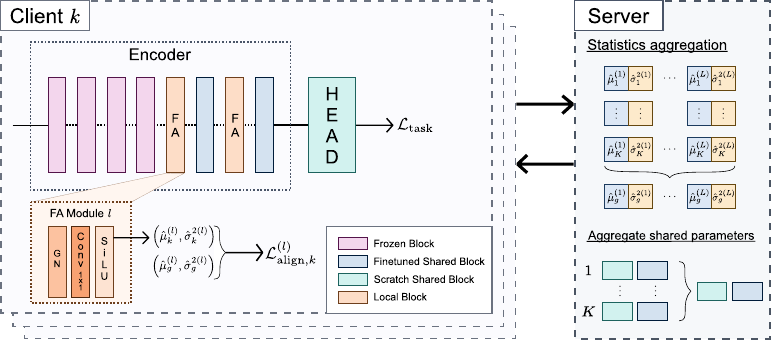}
    \caption{Overview of the Personalized Aligner Federated Pretrained (PAFP) framework}
    \label{fig:pipeline}
\end{figure*}

\section{Problem Formulation}
Let $K$ be the number of clients, where each client $k$ holds a private dataset $\mathcal{D}_k\coloneqq\{x_{k,i},y_{k,i}\}_{i=1}^{n_k}$ of size $n_k$. 
The total number of samples is denoted by $N\coloneqq \sum_{k=1}^Kn_k$.
Multicentric medical imaging is inherently exposed to multi-level data heterogeneity. Let $\mathbb{P}_k(\cdot,\cdot)$ and $\mathbb{P}_k(\cdot|\cdot)$ denote the joint and conditional data distributions for client $k$, respectively. We consider the following forms of statistical skew, which naturally arise in medical datasets:
\begin{itemize}[leftmargin=*]
    \item \textbf{Feature Skew} (covariate shift): $\mathbb{P}_i(X) \neq \mathbb{P}_j(X)$ while $\mathbb{P}_i(Y|X) = \mathbb{P}_j(Y|X)$, for all $i \neq j$. Variations in acquisition devices, practitioners, and clinical protocols result in inconsistent feature space distributions across medical centers.
    \item \textbf{Label Skew} (prior probability shift): $\mathbb{P}_i(Y) \neq \mathbb{P}_j(Y)$ while $\mathbb{P}_i(X|Y) = \mathbb{P}_j(X|Y)$, for all $i \neq j$. Geographical disparities and center-specific specializations lead to variations in label distributions across clients.
    \item \textbf{Concept Drift}: $\mathbb{P}_i(Y|X) \neq \mathbb{P}_j(Y|X)$ while $\mathbb{P}_i(X) = \mathbb{P}_j(X)$, for all $i \neq j$. Inter-observer variability and differing clinical guidelines cause the relationship between features and labels to vary across centers.
\end{itemize}
In addition to these distributional shifts, medical imaging datasets exhibit \textbf{quantity skew} ($n_i \neq n_j$ for clients $i \neq j$). Unlike the previous types, quantity skew is distribution-agnostic and can occur even when $\mathbb{P}_i(X,Y) = \mathbb{P}_j(X,Y)$: it simply reflects imbalances in 
patient volume 
or collection duration. In practice, it frequently co-occurs with distributional shifts, further compounding the overall \textbf{\textit{mixed skew}} challenge. 

To formally address these heterogeneities, we adopt a decoupled pFL paradigm. We formulate the optimization objective to jointly learn a set of shared global representations alongside client-specific parameters:
\begin{align}
&\min_{w, \{v_k\}_{k=1}^K} \sum_{k=1}^K \tfrac{n_k}{N} F_k(w, v_k), \\
&\text{with}~\,F_k(w,v_k) \coloneqq \mathbb{E}_{(x,y) \sim \mathcal{D}_k}\big[\mathcal{L}(g(x;w,v_k),y)\big],
\end{align}
where $\mathcal{L}$ denotes the loss function, and $g$ is a neural network with  global weights $w$ and local personalized weights $v_k$ for each client $k$.



Furthermore, medical imaging datasets typically feature a limited number of samples relative to their high-dimensional feature spaces 
(e.g., high-resolution volumetric or multi-modal scans). Training a model from scratch in this regime is  unrealistic, as it would require vastly more data than is available locally, or even globally. 
This sample-to-dimension deficit restricts the space of admissible global models $w$ to those leveraging pre-trained representations. 
In the following section, we present a method designed to jointly address the mixed skews under the pFL objective 
and the pre-trained backbone constraint defined above.





    




\section{Method}

This section describes the proposed PAFP framework for the classification task. We consider a neural network $g\coloneqq h\circ f$, where $f$ is a pre-trained backbone that maps from the input space $\mathcal{X}$ to a representation space $\mathcal{Z}$, and $h$ is a classifier head that maps from $\mathcal{Z}$ to the label space $\mathcal{Y}$.



The full training pipeline is summarized by Fig.~\ref{fig:pipeline}.

\subsection{Feature Alignment modules}

Feature skew induces distributional divergence not only in the input space $\mathcal{X}$, but also across the intermediate representation spaces $H^{(l)}$ inside the feature extractor. Specifically, for a pre-trained model where $L$ blocks are fine-tuned, the hidden representation at each fine-tuned layer $l$, denoted $H^{(l)}$, exhibits a divergence in its feature distribution. Left unaddressed, this divergence propagates through the fine-tuning process, causing model divergence and ultimately leading to a collapse in consensus between 
fine-tuned client layers.

To mitigate this, we introduce Feature Alignment (FA) modules, each inserted before a corresponding fine-tuned block. Each FA module consists of a Group Normalization layer, a position-wise affine transformation across channels implemented as a $1\times1$ convolution, and a SiLU activation, operating jointly as a learned transformation of the local feature distribution. 

The parameters of the FA modules are client-specific and belong to $v_k$. Their role is to realign each client's local representations toward a shared target distribution obtained by aggregating statistics across clients, so that every fine-tuned block receives a representation $H^{(l)}$ whose feature distribution is consistent across clients. This consistency limits the divergence of local updates and improves agreement on the shared weights $w$.

Our FA modules are inspired by FDSE~\cite{wang25dse}, which similarly introduces 
learnable modules to address distributional shift across clients. Our approach 
differs in two key aspects. First, we update the global target statistics using 
an Exponential Moving Average across communication rounds, rather than a direct 
average, in order to stabilize convergence and reduce sensitivity to round-to-round 
fluctuations in client participation. Second, since the FA modules are injected 
directly into a pre-trained backbone, we initialize the $1\times1$ convolution 
using a Dirac delta (identity) initialization~\cite{zagoruyko2017diracnets}, setting its 
kernel to an identity mapping with zero bias, while the Group Normalization 
affine parameters are initialized to unit scale and zero shift. This ensures 
that, at the start of fine-tuning, the FA modules do not perturb the pre-trained 
representations, thereby avoiding the training instability that would otherwise 
arise from inserting randomly initialized layers into an already-converged 
backbone.

\paragraph{Local distribution} 
To guide the FA modules toward a shared target distribution, we introduce an alignment loss that acts as a regularization between the local statistics at layer $l$ and the estimated global distribution.
Let $\mu_k^{(l)} \in \mathbb{R}^{C^{(l)}\times H^{(l)} \times W^{(l)}}$ and $\sigma_k^{2(l)} \in \mathbb{R}^{C^{(l)}\times H^{(l)} \times W^{(l)}}$ denote the spatial mean and variance, computed over the batch dimension, of the output of the $l$-th FA module for client $k$. We estimate the local running statistics over a complete epoch using Exponential Moving Average (EMA) with momentum $\beta_1$, such that:
\begin{equation}
    \begin{aligned}
    \hat{\mu}_k^{(l)}      &\leftarrow (1-\beta_1)\, \hat{\mu}_k^{(l)} + \beta_1 \, \mu_k^{(l)},     \\
    \hat{\sigma}_k^{2(l)}  &\leftarrow(1-\beta_1)\, \hat{\sigma}_k^{2(l)} + \beta_1 \, \sigma_k^{2(l)}.
\end{aligned}
\end{equation}

\paragraph{Global target distribution}
At the end of each communication round $t$, each client transmits its local 
statistics $\{\hat{\mu}_k^{(l)}, \hat{\sigma}_k^{2(l)}\}_{l=1}^L$ to the server alongside its locally fine-tuned global weights ($w_k^t$). The server estimates the global target distribution by aggregating the local statistics received from all clients:
\begin{equation}
    \mu_{g}^{(l)} \coloneqq \sum_{k=1}^K \frac{n_k}{N} \hat{\mu}_k^{(l)}, \quad
    \sigma_{g}^{2(l)} \coloneqq \sum_{k=1}^K \frac{n_k}{N} \hat{\sigma}_k^{2(l)}.
\end{equation}
To ensure stability across communication rounds, the global target statistics are smoothed via EMA with momentum $\beta_2$:
\begin{equation}
\begin{aligned}
    \hat{\mu}_g^{(l)}&\leftarrow (1-\beta_2)\, \hat{\mu}_g^{(l)}      + \beta_2\, \mu_{g}^{(l)},\\
    \hat{\sigma}_g^{2(l)}&\leftarrow (1-\beta_2)\, \hat{\sigma}_g^{2(l)}  + \beta_2\, \sigma_{g}^{2(l)}.
    \end{aligned}
\end{equation}

\paragraph{Alignment loss}
Each client computes the following alignment loss to guide its FA modules to transform local feature distributions toward the estimated global target:
\begin{align}
    \mathcal{L}_{\text{align},k} &\coloneqq \sum_{l=1}^{L} \mathcal{L}_{\text{align},k}^{(l)}, \\
    \text{with}~\ \mathcal{L}_{\text{align},k}^{(l)} &\coloneqq \text{MSE}\!\left(\hat{\mu}_k^{(l)},\, \hat{\mu}_g^{(l)}\right) + 
        \text{MSE}\!\left(\hat{\sigma}_k^{2(l)},\, \hat{\sigma}_g^{2(l)}\right).
\end{align}
The total local objective for client $k$ combines the task loss 
$\mathcal{L}_{\text{task},k}$ with the alignment loss as a regularization term:
\begin{equation}
    \mathcal{L}_k(w, v_k) \coloneqq \mathcal{L}_{\text{task},k}(w, v_k) + 
    \lambda \cdot \mathcal{L}_{\text{align},k}(v_k),
\end{equation}
where $\lambda > 0$ controls the regularization strength.

\begin{figure}[!t]
\begin{algorithmic}[1]
\REQUIRE global model $w$, local epochs $E$, rounds $T$
\FOR{each round $t=0,1\dots,T$}
    \STATE Server sends $\{ w; ( \hat{\mu}_g, \hat{\sigma}_g^2) \}$ to each client $k$
    \FOR{each client $k \in [K]$ in parallel}
        \FOR{each epochs}
            \FOR{each batch $b$}
                \STATE $\hat{\mu}_{k,b}\gets(1-\beta_1)\cdot\hat{\mu}_{k,b-1}+\beta_1 \cdot \mu_{k,b}$
            \ENDFOR
        \ENDFOR
        \IF{alternate case}
            \IF{local round}
            \STATE Compute loss $\mathcal{L}_k \leftarrow \mathcal{L}_{\text{task},k}+\lambda \mathcal{L}_{\text{align},k}$
            \STATE ${v_k^t} \gets \text{SGD}(\mathcal{L}_k)$ for $E$ epochs
            \STATE $\{(\hat{\mu}_{k}, \hat{\sigma}^2_{k}) \}$ send to server
            \ENDIF
            
            \IF{global round}
            \STATE Compute loss $\mathcal{L}_k \leftarrow \mathcal{L}_{\text{task},k}$
            \STATE ${w_k^t} \gets \text{SGD}(\mathcal{L}_k)$ for $E$ epochs
            \STATE $\{w_k^t, (\hat{\mu}_{k}, \hat{\sigma}^2_{k}) \}$ send to server
            \ENDIF
        \ENDIF
        \IF{non-alternate case}
            \STATE Compute loss $\mathcal{L}_k \leftarrow \mathcal{L}_{\text{task},k}+\lambda \mathcal{L}_{\text{align},k}$
        \ENDIF
    \ENDFOR
    \STATE $w^{t+1} \gets \sum_{k=1}^K \frac{n_k}{N}w_k^t$
    \STATE $(\mu_{\text{g,round}},\sigma_{\text{g,round}}^2)\gets\sum_{k=1}^K\frac{n_k}{N}(\hat{\mu}_{k}, \hat{\sigma}^2_{k})$
    \STATE $(\hat{\mu}_{\text{g}},\hat{\sigma}_{\text{g}}^2)=(1-\beta_2)\cdot(\hat{\mu}_{\text{g}},\hat{\sigma}_{\text{g}}^2)+\beta_2\cdot(\mu_{\text{g,round}}, \sigma_{\text{g,round}}^2)$
\ENDFOR
\end{algorithmic}
\caption{Personalized Aligner Federated Pretrained Algorithm.\label{algo:training}}
\end{figure}

\subsection{Optimization Strategy}
We introduce two variants of our method. Standard PAFP uses a single optimizer to update both the local parameters $v_k$ and shared parameters $w$ based on the joint loss. Alternatively, PAFP-alt employs two optimizers: one updates $w$ using only the task loss, and another updates $v_k$ using both the task and alignment losses. This variant alternates between optimizing shared and local parameters, with a hyper-parameter $r$ dictating that local updates occur every $r$ rounds. The full training procedure, including client-side local updates and server-side aggregation, is detailed in Fig.~\ref{algo:training}.

\subsection{Logits correction}

Label skew induces client-specific imbalances in the label distribution 
$\mathbb{P}_k(Y)$, which biases the classifier head $h$ toward the label 
distribution observed locally during fine-tuning. Left unaddressed, this bias 
causes the model's output logits to systematically favor over-represented 
classes at each client, degrading performance on under-represented classes.

To mitigate this, in the experiments, we tested a Logit Corrector (LC), a lightweight, client-specific 
adaptation layer inserted immediately after the classifier head $h$. The Logit 
Corrector applies an elementwise affine transformation to the output logit 
vector, allowing each client to shift and scale its predictions to better match 
its local label distribution. Formally, let $\hat{y}_k = h(z; w^h) \in 
\mathbb{R}^C$ denote the raw logit vector produced by the classifier head for 
client $k$, where $C$ is the number of classes. The corrected logit vector is 
given by:
\begin{equation}
    \hat{y}'_k \coloneqq a_k \odot \hat{y}_k + b_k
\end{equation}
where $a_k \in \mathbb{R}^C$ and $b_k \in \mathbb{R}^C$ are client-specific 
scaling and shifting vectors, respectively, and $\odot$ denotes the Hadamard product.

As with the FA modules, the parameters $\{a_k, b_k\}$ of the Logit 
Corrector are client-specific and constitute part of $v_k$.

\section{Experimental setup}

\paragraph{Dataset} We evaluate our method on FedISIC, a federated skin lesion 
classification dataset from the FLamby benchmark~\cite{NEURIPS2022_232eee8e}, derived from the 
ISIC 2019/2020 challenges. FedISIC partitions dermoscopic images across $K$ 
clients according to their original acquisition center, providing a naturally 
occurring, non-simulated instance of multicentric data heterogeneity.

\paragraph{Baseline model and training} Following the FedISIC baseline, we use 
an EfficientNet backbone pre-trained on ImageNet as the feature extractor $f$, 
followed by a linear classification head $h$. As the task loss $\mathcal{L}_{\text{task},k}$, 
we reuse the baseline focal loss~\cite{lin2017focal} with class weights $\alpha_k \in \mathbb{R}^C$ 
computed from the local label distribution of each client $k$, and a fixed 
focusing parameter $\gamma = 2.0$.
For the feature alignment modules, we set the regularization strength to $\lambda = 10$, and the momentum parameters to $\beta_1 = 0.8$ for the local modules and $\beta_2 = 0.8$ for the server side. 

All clients are trained using an SGD optimizer with a learning rate of $10^{-2}$ and a weight decay of $10^{-5}$. Training is conducted for up to $1500$ communication rounds—though most methods converge well before this limit—with each client performing $1$ local epoch per round using a batch size of $64$.

\begin{table}[!t]
    \centering
    \caption{FedISIC Client Label Distribution}\label{tab:fedisic_label}
    
        \centering
        \begin{tabular}{c|cccccccc}
            \hline
            & \multicolumn{8}{c}{Class} \\ \cline{2-9}
            Client & $C_0$ & $C_1$ & $C_2$ & $C_3$ & $C_4$ & $C_5$ & $C_6$ & $C_7$\\
            \hline
            1 (n=12413) & 22.6\% & 34.0\% & 22.7\% & 6.1\% & 9.3\% & 1.0\% & 0.9\% & 3.3\%\\
            2 (n=3954) & 0.6\% & 94.1\% & 0.1\% & 0.00\% & 3.0\% & 0.7\% & 1.4\% & 0.00\%\\
            3 (n=3359) & 19.8\% & 54.6\% & 6.5\% & 0.7\% & 14.1\% & 1.5\% & 2.4\% & 0.4\%\\
            4 (n=2258) & 15.3\% & 35.9\% & 12.3\% & 4.8\% & 22.5\% & 1.1\% & 0.2\% & 8.0\%\\
            5 (n=814) & 25.3\% & 53.3\% & 0.0\% & 0.0\% & 21.4\% & 0.0\% & 0.0\% & 0.0\%\\
            6 (n=437) & 16.5\% & 78.6\% & 1.4\% & 0.00\% & 1.7\% & 1.1\% & 0.6\% & 0.00\%\\

            \hline
        \end{tabular}

\end{table}

\begin{figure}[!t]
  \centering
  \includegraphics[width=\linewidth]{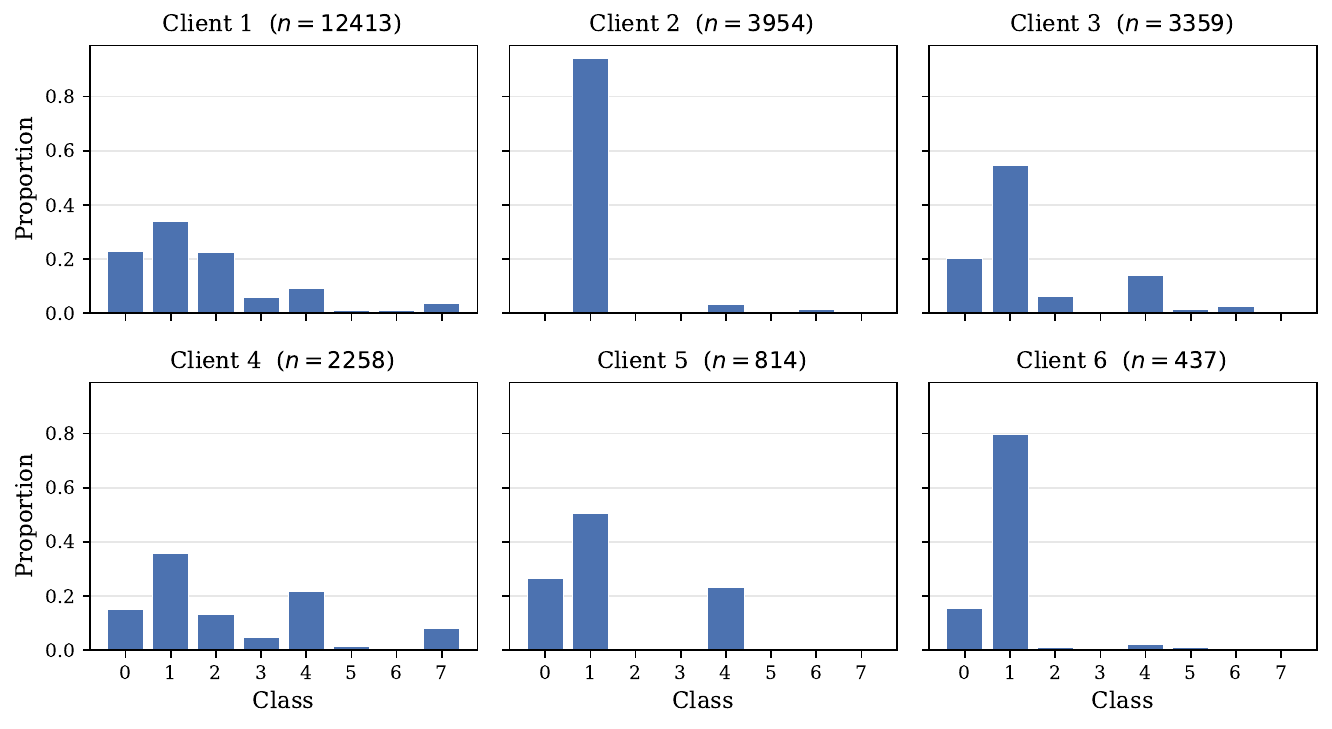}
  \caption{Per-client data distributions across the six clients.}
  \label{fig:client-dist}
\end{figure}
\begin{figure}[!t]
  \centering
  \includegraphics[width=.8\linewidth]{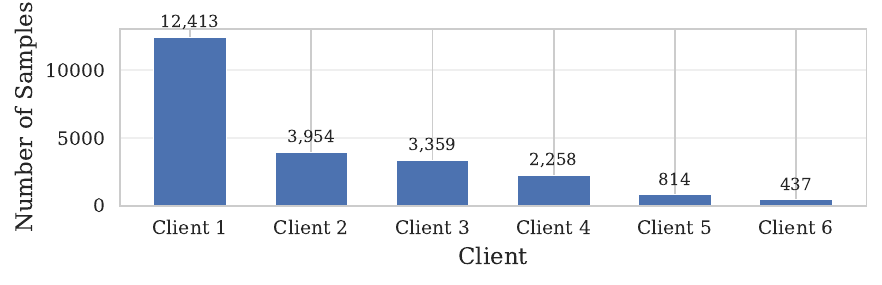}
  \caption{Client size distribution.}
  \label{fig:client-quantity}
\end{figure}

\begin{figure}[!t]
  \centering
  \includegraphics[width=0.80\linewidth]{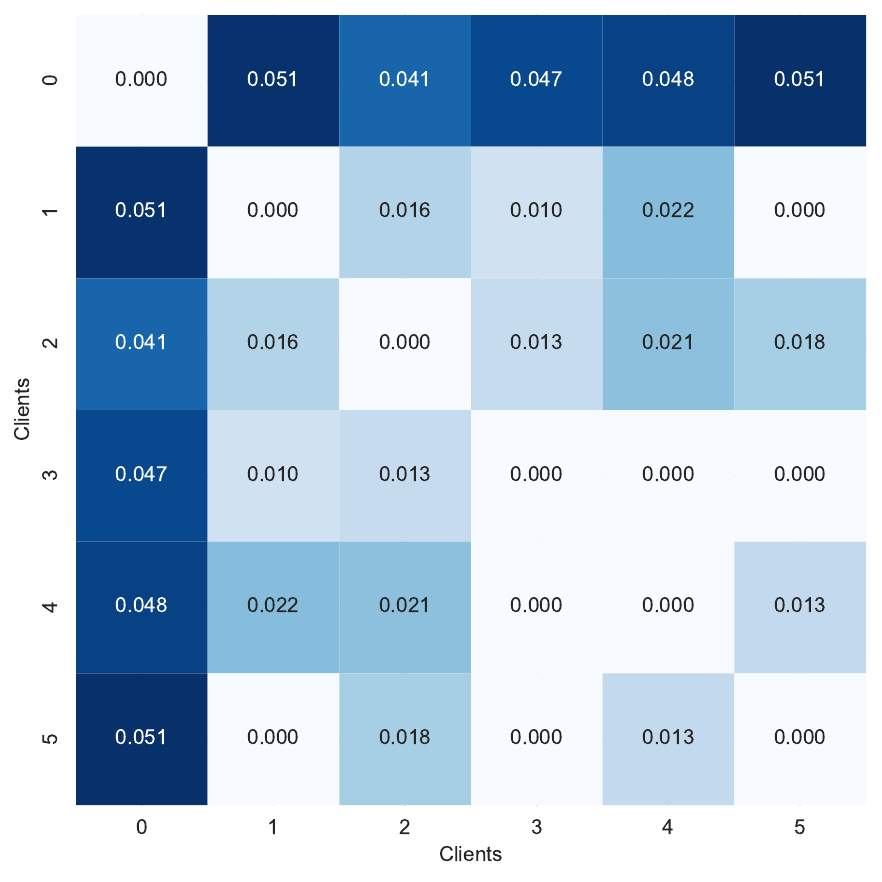}
  \caption{FedISIC interclient MMD}
  \label{fig:fedisic_mmd}
\end{figure}

\paragraph{Characterizing dataset skew} We first characterize the heterogeneity 
present in FedISIC prior to training. Table~\ref{tab:fedisic_label} reports the 
per-client label distribution, expressed as the percentage of samples belonging 
to each class, confirming the presence of label skew across clients. Fig.~\ref{fig:client-dist} depicts the label distribution of each client and Fig.~\ref{fig:client-quantity} highlight quantity skew. To quantify 
feature skew, we compute the pairwise Maximum Mean Discrepancy (MMD) between 
clients' latent representations $Z$, extracted from the pre-trained backbone 
$f$ prior to any fine-tuning. We use a Multiple RBF Kernel with $\sigma$ ranging from a scaling of $[10^{-3}, 10^{-2}, 10^{-1}, \ 1, 10^{1}, 10^{2}, 10^{3}]$ apply to a base gamma obtained using the median heuristic.
Table~\ref{fig:fedisic_mmd} reports  the resulting pairwise MMD matrix, quantifying the initial feature-space shift between clients before federated training begins.
As commonly observed in the literature, empirical pairwise MMD estimates on latent representations may violate the triangle inequality due to finite-sample estimation noise, yet they reliably capture the relative statistical discrepancies across clients.

\begin{table*}[!t]
    \centering
    \caption{Balanced Accuracy on FedISIC (mean$\pm$std, best federated in bold, second best federated in italic)}\label{tab:fedisic_result}
    
    \centering
    \begin{tabular}{l|cccccc|cc|c}
        \hline
        & client 1 & client 2 & client 3 & client 4 & client 5 & client 6 & weight avg & unweighted avg & inter-client variance\\
        & (N=12413) & (N=3954) & (N=3359) & (N=2258) & (N=814) & (N=437) & & & \\
        \hline
        isolated & $80.16\pm1.01$ & $75.45\pm5.30$ & $70.69\pm0.89$ & $70.20\pm4.59$ & $64.17\pm1.85$ & $35.67\pm2.99$ & $75.62\pm1.69$ & $66.06\pm1.88$ & $2.16\pm0.46$ \\ 
        \hline
        
        FedAVG ft & $\textbf{79.99}\pm\textbf{0.95}$ & $55.94\pm6.86$ & $\textit{67.10}\pm\textit{5.63}$ & $57.47\pm3.84$ & $23.56\pm1.08$ & $55.57\pm5.46$ & $69.39\pm1.70$ & $56.60\pm1.92$ & $3.09\pm0.28$ \\ 
        
        DBE ft & $\textit{79.86}\pm\textit{0.79}$ & $60.26\pm6.94$ & $\textbf{68.97}\pm\textbf{7.37}$ & $59.29\pm6.09$ & $35.68\pm5.49$ & $\textbf{66.85}\pm\textbf{3.42}$ & $71.15\pm1.69$ & $61.82\pm1.87$ & $2.09\pm0.59$ \\ 
        
        DITTO ft & $78.97\pm1.25$ & $51.40\pm4.70$ & $65.05\pm10.36$ & $61.21\pm3.40$ & $30.84\pm4.52$ & $45.55\pm6.65$ & $68.21\pm2.26$ & $55.50\pm2.96$ & $2.12\pm0.56$ \\ 
        \hline
        
        PAFP & $76.49\pm0.73$ & $\textbf{76.52}\pm\textbf{5.49}$ & $64.92\pm0.89$ & $\textit{66.24}\pm\textit{1.31}$ & $\textit{64.83}\pm\textit{0.74}$ & $53.23\pm14.44$ & $\textbf{72.97}\pm\textbf{1.34}$ & $\textit{67.04}\pm\textit{2.51}$ & $0.97\pm0.76$ \\
        
        PAFP-alt (r=10) & $76.53\pm1.51$ & $71.88\pm5.99$ & $64.38\pm4.48$ & $66.19\pm1.12$ & $63.62\pm1.46$ & $\textit{62.56}\pm\textit{2.30}$ & $\textit{72.26}\pm\textit{1.42}$ & $\textbf{67.53}\pm\textbf{1.36}$ & $\textbf{0.35}\pm\textbf{0.13}$ \\ 
        
        \hline
        
        ft 2B & $78.48\pm0.69$ & $69.02\pm2.42$ & $63.08\pm5.12$ & $65.94\pm1.10$ & $59.72\pm2.48$ & $37.53\pm5.00$ & $71.99\pm0.61$ & $62.29\pm0.42$ & $1.67\pm0.42$ \\
        
        ft 2B + LC & $77.98\pm0.94$ & $67.53\pm7.18$ & $64.55\pm4.71$ & $65.03\pm2.12$ & $57.79\pm1.20$ & $45.07\pm12.77$ & $71.67\pm1.59$ & $62.99\pm1.92$ & $1.37\pm0.63$ \\
        
        PAFP + LC & $76.40\pm0.47$ & $\textit{73.25}\pm\textit{7.45}$ & $65.42\pm0.94$ & $\textbf{66.31}\pm\textbf{0.57}$ & $63.05\pm1.91$ & $40.09\pm11.25$ & $72.14\pm1.45$ & $64.09\pm2.13$ & $1.63\pm0.77$ \\
        
        PAFP-alt (r=50) & $76.46\pm0.85$ & $71.82\pm9.70$ & $64.45\pm4.04$ & $63.93\pm1.39$ & $\textbf{64.92}\pm\textbf{1.12}$ & $36.72\pm3.67$ & $71.56\pm2.19$ & $63.05\pm1.80$ & $1.41\pm0.36$ \\ 
        
        
        PAFP-alt (r=2) & $76.75\pm0.26$ & $69.96\pm6.00$ & $65.60\pm0.73$ & $65.41\pm2.08$ & $62.60\pm1.79$ & $51.72\pm13.45$ & $71.91\pm1.12$ & $65.34\pm2.03$ & $\textit{0.91}\pm\textit{0.57}$ \\ 
        \hline
    \end{tabular}

\end{table*}
\section{Results}

\paragraph{Evaluation protocol} For each experiment, we report the mean and standard deviation of the balanced accuracy over 5 independent runs.
We evaluate all baselines from a pFL perspective, even those that were not originally designed as pFL methods.
Concretely, for every method we report the performance of each client's local fine-tuned model at the end of its local training, rather than that of the aggregated global model as is conventional for traditional FL methods.
This ensures all methods are assessed on the same footing, namely their per-client personalized performance. 
We report  per-client balanced accuracy, as well as two aggregate measures:
the weighted  average, using data proportion $n_k/N$, and the unweighted average, using equal  client weighting $1/K$, to assess overall system performance from both a population-level and a client-level perspective, respectively.
In the context of pFL where the goal is to obtain the best local model for each client, an unweighted average metric provides a more faithful assessment of the system, as it prevents large clients from masking the poor performance of smaller ones.
To evaluate whether a given method reduces performance disparity across clients, we further report the inter-client variance of balanced accuracy, computed within each run and then averaged (with standard deviation) across the 5 runs.

Experimental results are presented in Table~\ref{tab:fedisic_result}. The isolated case corresponds to a non-FL scenario where each client trains a model solely on its own local data. These isolated results are included solely as a baseline reference, with our analysis focused primarily on the federated training outcomes. We benchmark the FL baselines FedAvg, DBE, and Ditto in a fine-tuned version (ft), since we start from a pre-trained model, which departs from the train-from-scratch assumption these methods were designed for.
We deliberately exclude Parameter-Efficient Fine-Tuning (PEFT) methods from our evaluation. While they share similarities with our local adaptation objective, our primary focus is isolating and addressing data skew rather than optimizing parameter efficiency.

We present PAFP and PAFP-alt ($r=10$) as the best-performing configurations, and ft-2B, ft-2B + LC, PAFP + LC, PAFP-alt ($r=50$), and PAFP-alt ($r=2$) as exploratory experiments investigating various configuration variations. Here, ``ft-2B`` denotes the EfficientNet model fine-tuned exclusively on its last two blocks, and ``LC'' indicates the incorporation of the Logit Corrector.

Overall, we observe that the PAFP variants achieve the best overall performance under both evaluation schemes. The alternate variant (PAFP-alt) obtains the highest unweighted average balanced accuracy ($67.53\pm1.36$\%) and the second-highest weighted average ($72.26\pm1.42$\%). The highest weighted average is attained by non-alternate PAFP ($72.97\pm1.34$\%).
The best per-client results for clients 2, 4, and 5 are likewise attained by PAFP configurations (PAFP, PAFP + LC, and PAFP-alt ($r=50$), reaching $76.52\pm5.49$\%, $66.31\pm0.57$\%, and $64.92\pm1.12$\%, respectively). Furthermore, PAFP-alt ($r=10$) reduces the inter-client variance of balanced accuracy to $0.35$ (compared to the lowest baseline variance of $2.09$). Thus, the configuration that is most accurate under the client-level unweighted average is also the most equitable across clients, rather than trading one for the other.

Results from the isolated (non-FL) setting show that clients holding large amounts of data benefit little from federation. This is most visible for client 1 ($80.16\pm1.01$\% isolated vs. $79.99\pm0.95$\% FedAvg-ft), whose performance is already close to its local-only ceiling.

Finally, by comparing each configuration utilizing an LC module (PAFP + LC and ft-2B + LC) to its counterpart without LC (PAFP and ft-2B), we note that adding the Logit Corrector does not improve performance.

\section{Discussion}

\paragraph{Interpreting the main result} The strongest configuration couples the
FA modules with an alternating optimization schedule, and the gap between this
variant and the single-optimizer one is informative. Because PAFP fine-tunes a
backbone pre-trained on natural images, all clients start from a common,
already-structured latent space. Before inter-client heterogeneity becomes the
dominant obstacle, every client must first adapt the shared weights $w$ from the
natural-image domain to the medical-imaging domain. Only once $w$ has specialized
to medical imaging does the residual divergence between client distributions
become the problem the FA modules are meant to solve. We hypothesize that alternating between rounds that update the shared parameters and rounds that update the local parameters allows the system to interleave these two processes: global domain adaptation (natural to
medical) and local adaptation to each client's distribution. In the
single-optimizer variant, by contrast, the alignment gradient flows into $w$ from
the first round, pushing the shared backbone to satisfy a client-specific
alignment objective before it has adapted to the medical domain. This conflates
the two adaptation problems and perturbs the shared representation, which we
believe explains its lower and less stable performance. That the alternating
variant simultaneously attains the best unweighted average and the lowest
inter-client variance supports this reading: shielding $w$ from the alignment
gradient yields both a better shared representation and more equitable
personalization.

\paragraph{Limitations} Several limitations temper these conclusions. First, our evaluation covers a single dataset and imaging modality (FedISIC dermoscopy). While the problem formulation and the PAFP architecture make no domain-specific assumptions, the generality of the approach beyond dermoscopic classification remains argued rather than empirically demonstrated. Second, although we formalize four types of skew, only feature skew and label skew receive dedicated mechanisms. Concept drift and quantity skew are handled only implicitly: quantity skew in particular is reflected in our unweighted evaluation and in the data-proportional aggregation, but PAFP contains no component that explicitly targets it. Third, the FA modules introduce a modest communication overhead, since each client transmits per-layer feature statistics alongside its weights at every round. This corresponds to 56,672 statistics, equivalent to an increase of 453 kB of communication per client. Compared to the 3,181,612 shared parameters, this represents a mere 1.7\% increase in communication volume and can thus be considered negligible. Finally, because the backbone model was pretrained on natural images rather than medical imaging, the observed performance gains may be confounded by the domain shift. Consequently, further experiments are required to isolate whether the proposed method effectively mitigates client heterogeneity independent of this initial domain gap.

\paragraph{Perspectives} Three directions follow naturally from this work. The most immediate is to extend PAFP to Magnetic Resonance Imaging (MRI) , a non-quantitative modality whose signal intensities are not standardized across scanners, field strengths, and acquisition protocols, and which therefore exhibits particularly strong and structured inter-client feature skew. The second direction is to leverage a medical imaging foundation model to mitigate the domain gap that the backbone must bridge during early training, thereby isolating data skew as the primary bottleneck and more rigorously demonstrating PAFP's efficacy.
The third direction is to address quantity skew explicitly within the mixed-skew setting, rather than leaving it to the aggregation and evaluation scheme, so that data-scarce clients are supported by a dedicated mechanism as feature and label skew already are. Together, these extensions would move the framework from a single-modality proof of concept toward the general treatment of mixed skew that motivates it.







\bibliographystyle{IEEEtran}
\bibliography{IEEEabrv,mybibfile}

\end{document}